\documentclass{article}

 \PassOptionsToPackage{numbers, compress}{natbib}

 \usepackage[preprint]{nips_2018}

\usepackage[utf8]{inputenc} 
\usepackage[T1]{fontenc}    
\usepackage{hyperref}       
\usepackage{url}            
\usepackage{booktabs}       
\usepackage{amsfonts}       
\usepackage{nicefrac}       
\usepackage{microtype}      
\usepackage{graphicx}
\usepackage{verbatim}
\usepackage{subcaption}
\usepackage{multirow}
\usepackage{enumitem}
\setlist[enumerate]{itemsep=0mm}
\setlist[itemize]{itemsep=0mm}

\title{Machine learning models for prediction of droplet collision outcomes}

\author{
  Arpit Agarwal \\
  \texttt{arpit.r.agarwal@gmail.com} \\
}

\begin{document}

\maketitle

\begin{abstract}
Predicting the outcome of liquid droplet collisions is an extensively studied phenomenon but the current physics based models for predicting the outcomes are poor (accuracy $\approx 43\%$). The key weakness of these models is their limited complexity. They only account for 3 features while there are many more relevant features that go unaccounted for. This limitation of traditional models can be easily overcome through machine learning modeling of the problem. In an ML setting this problem directly translates to a classification problem with 4 classes. Here we compile a large labelled dataset and tune different ML classifiers over this dataset. We evaluate the accuracy and robustness of the classifiers. ML classifiers, with accuracies over 90\%, significantly outperform the physics based models. Another key question we try to answer in this paper is whether existing knowledge of the physics based models can be exploited to boost the accuracy of the ML classifiers. We find that while this knowledge improves the accuracy marginally for small datasets, it does not improve accuracy with if larger datasets are used for training the models.
\end{abstract}

\section{Introduction}
\subsection{Motivation}
Droplet collisions are important to the dynamics of a liquid spray. Sprays of engineering relevance (e.g., automotive, food processing, printing) have a very high droplet density and therefore a lot of droplet collisions. For example, diesel sprays have been reported to have 10$^8$ collisions per cm$^3$ per microsecond~\cite{macinnes1991comparisons}. Collisions have a direct effect on the liquid drops as they change the droplet sizes and velocities. Due to the high frequency of droplet collisions, and their cascading effect on the system dynamics, it is important to accurately model the droplet collision phenomena.

The problem of modeling the outcome of droplet collisions has been worked on extensively by fluid dynamicists. The models proposed in literature for predicting the outcomes of droplet collisions are phenomenological. They are based simplified analyses of the governing fluid dynamics equations. \citet{agarwal2019computational} evaluate the current state of the art in physics based models and report an accuracy of around 64\% on the data they considered. On considering additional data reported by \citet{sommerfeld2016modelling} this accuracy drops to around 43\%. The complexity of these models is limited by our understanding of fluid mechanics, and our cognitive limitations. Models of higher complexity are needed to improve the accuracy.

The problem of droplet collision prediction can be directly framed as a 4-class classification problem. This problem is ideal for a machine learning solution because of two reasons:
\begin{enumerate}
\item Ground truth complexity is high and beyond physical modeling
\item The problem is well studied, hence a lot of experimental data has been published over the years
\end{enumerate}
Here I compile labelled datasets from literature (7898 points in all, see Table~\ref{tab:data}) and train machine learning classification models over this data. The entire dataset, source code for the different models and tests, and all the figures that appear in the paper are available on Github~\cite{githubrepo}.

\subsection{Fluid Mechanics Background}
\label{sec:background}
\begin{figure}[h]
	\centering
	\includegraphics[width=0.8\textwidth]{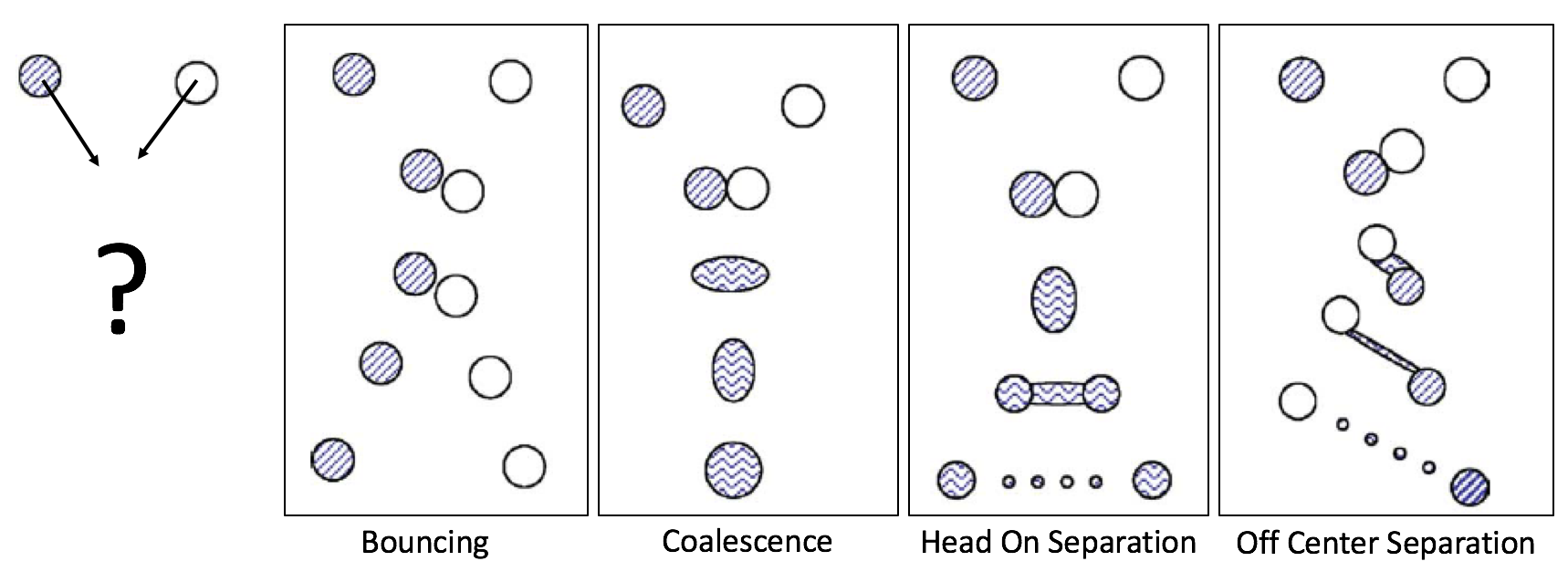}
	\caption{An illustration of the 4 types of outcomes expected - bouncing, coalescence, head-on separation (or reflexive separation), and off-center separation (or stretching separation). These are the classification labels for the present study.}
	\label{fig:outcomes}
\end{figure}

Four main types of outcomes are expected from collisions of two droplets, these are illustrated in Figure~\ref{fig:outcomes}. As this is a deterministic phenomenon, a priori prediction of collision outcomes is possible with enough information about the droplet and flow characteristics. Due to limitations of the techniques used in current fluid mechanics literature, only three main non-dimensional quantities, $\Delta, We$ and $B$ are considered in the outcome of a binary collision. It is well known that more features are relevant, but no models exist that account for all relevant features.

As highlighed in Figure~\ref{fig:outcomes} the four class labels (collision outcomes) are:
\begin{enumerate}
\item Coalescence
\item Bouncing
\item Stretching separation (or off-center separation)
\item Reflexive separation (or head-on separation)
\end{enumerate}

\begin{figure}[h!]
    \centering
    \begin{subfigure}[b]{0.45\textwidth}
        \includegraphics[width=\textwidth]{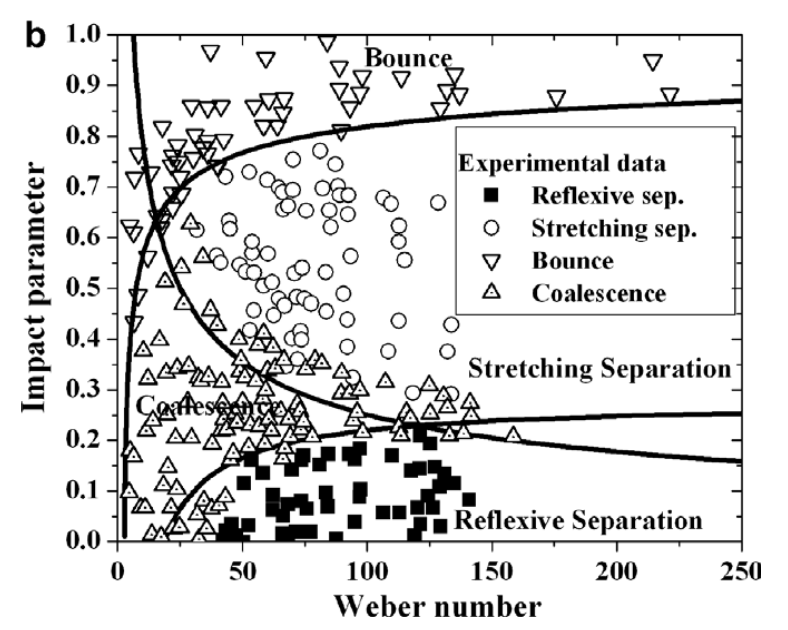}
        \caption{A sample dataset with the 4 labels and the physics based classification curves}
        \label{fig:munnannur_a}
    \end{subfigure}
    ~ 
    \begin{subfigure}[b]{0.45\textwidth}
        \includegraphics[width=\textwidth]{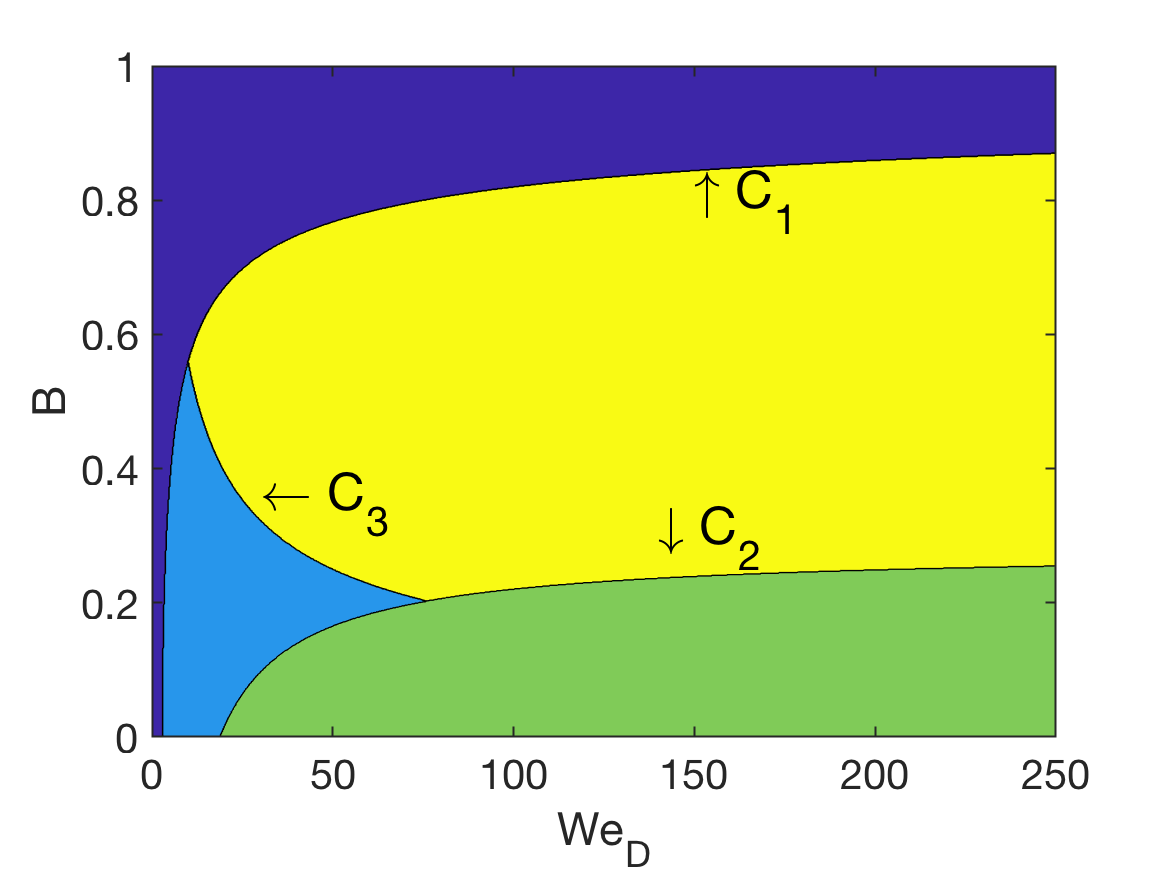}
        \caption{Classification curves \& map for the physics based models. The four different colors correspond to the four different labels.}
        \label{fig:regime}
    \end{subfigure}
    \caption{A sample dataset by \citet{munnannur2007new} and the physics based classification map by \citet{agarwal2019computational}.}
    \label{fig:munnannur}
\end{figure}

As per the traditional modeling approach, the 3-dimensional feature space is divided into 4 regions corresponding to the respective outcomes. Figure~\ref{fig:munnannur} shows a slice of this 3-dimensional space ($\Delta$ is set to 1), showing the four different outcomes. This figure also gives an example of a typical labelled experimental dataset.


\clearpage
\section{Dataset}

Data has been collected from four published articles. Details of the references are provided in Table~\ref{tab:data}. I manually extracted 7898 data points from the sources listed in Table~\ref{tab:data} using a web-based tool~\cite{rohatgi2011webplotdigitizer}. Each data point corresponds to one collision experiment.

\begin{table}[h]
  \caption{Source datasets}
  \label{tab:data}
  \centering
  \begin{tabular}{c l c p{5cm}}
    \toprule
    Date  & Source reference      				& No. of data points  & Comments\\
    \midrule
    1990 & \citet{ashgriz1990coalescence}     		& 986   &  \multirow{3}{5cm}{Early work - only basic features considered} \\
    1999 & \citet{estrade1999experimental} 		& 905   & \\ 
    1997 & \citet{qian1997regimes}    			& 217   & \\
    2016 & \citet{sommerfeld2016modelling}  		& 5790 & Newer work - emphasis on features that are not represented in traditional models\\
    \cmidrule(r){3-3}
             & 									& Total: 7898 & \\
    \bottomrule
  \end{tabular}
\end{table}

\begin{figure}[h]
	\centering
	\includegraphics[width=0.7\textwidth]{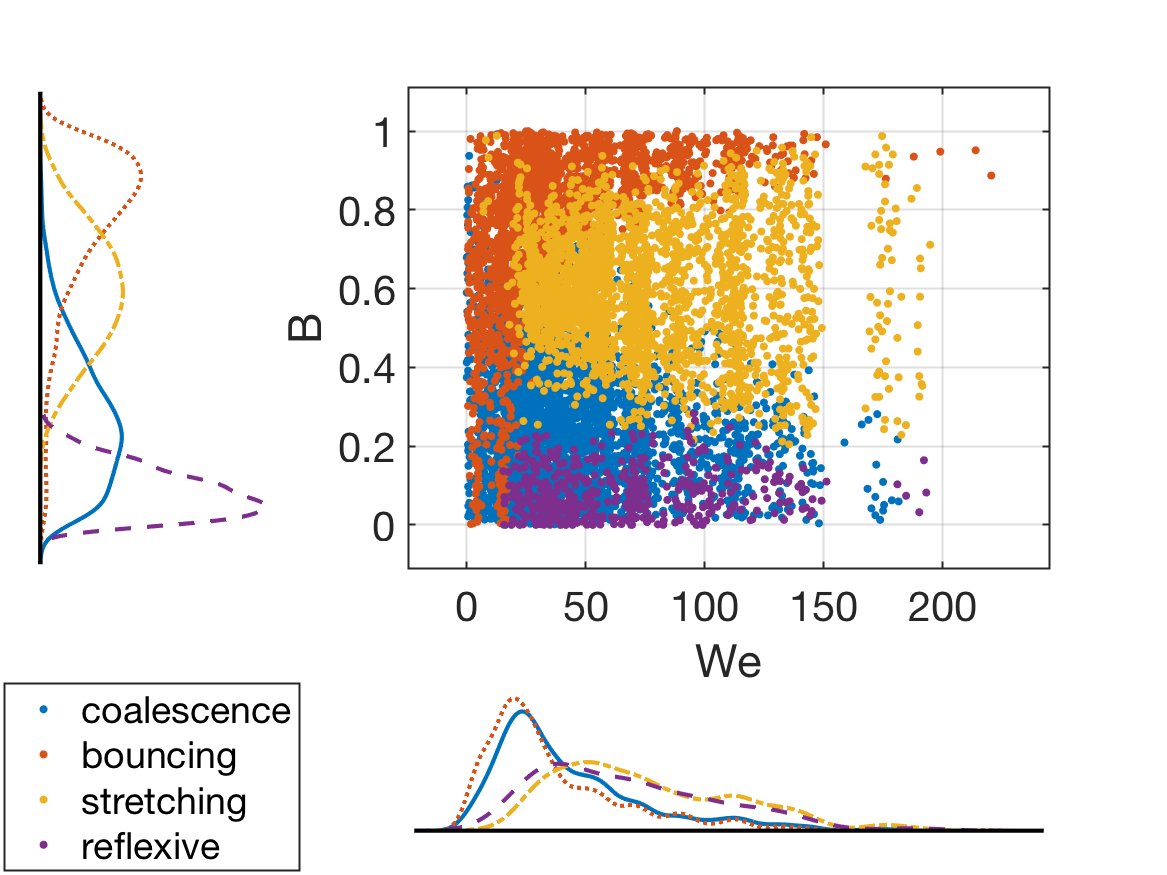}
	\caption{Overall droplet collision dataset (N = 7898, details in Table~\ref{tab:data}), showing variation in collision outcomes with respect to $We$ and $B$ (features 1 and 2).}
	\label{fig:data0}
\end{figure}

All the input features are numerical. The 5 base features and 3 additional features are described in Table~\ref{tab:features}. A slice of the overall dataset showing variation along $We$, and $B$ is presented in Figure~\ref{fig:data0}. Spray droplet collisions are most studied in the context of fuel sprays, and in typical physical operating ranges, $We$ and $B$ are the most relevant features in the outcome of the droplet collision. The distribution of the datapoints along the $We$ and $B$ axis are presented alongside the scatterplot.

\begin{table}[h]
  \caption{Description of the features}
  \label{tab:features}
  \centering
  \begin{tabular}{l l c p{5cm}}
    \toprule
    & 							Features		  			& Range & Comments\\
    \midrule
   \multirow{ 5}{*}{Base Features} 	& 1. $We$ (Weber number)      	& $\rm I\!R^+$ 	& See Figure~\ref{fig:data0}  \\
     							& 2. $B$ (Impact param.)		& [0, 1]  			& See Figure~\ref{fig:data0} \\
    							& 3. $\Delta$ (Drop size ratio)  	& [0, 1]  			& Most data has $\Delta = 1$; a few points with $\Delta = \{0.5, 0.75\}$ (see Figure~\ref{fig:data1})   \\
    							& 4. $P$ (Gas pressure) 		& $\rm I\!R^+$	&  Only discrete training data available (see Figure~\ref{fig:data2})\\
    							& 5. $\mu$ (Liquid viscosity) 		& $\rm I\!R^+$ 	& Only discrete training data available (see Figure~\ref{fig:data3}) \\
   \midrule
   \multirow{ 3}{3cm}{Additional Features (combinations of base features)} 	& Curve $C_1$     			& $\rm I\!R$ &  see eq.~\ref{eq:c1}\\
     							& Curve $C_2$					& $\rm I\!R$  & see eq.~\ref{eq:c2} \\
    							& Curve $C_3$  					& $\rm I\!R$  & see eq.~\ref{eq:c3}\\
    \bottomrule
  \end{tabular}
\end{table}

\begin{figure}[h!]
    \centering
    \begin{subfigure}[b]{0.4\textwidth}
        \includegraphics[width=\textwidth]{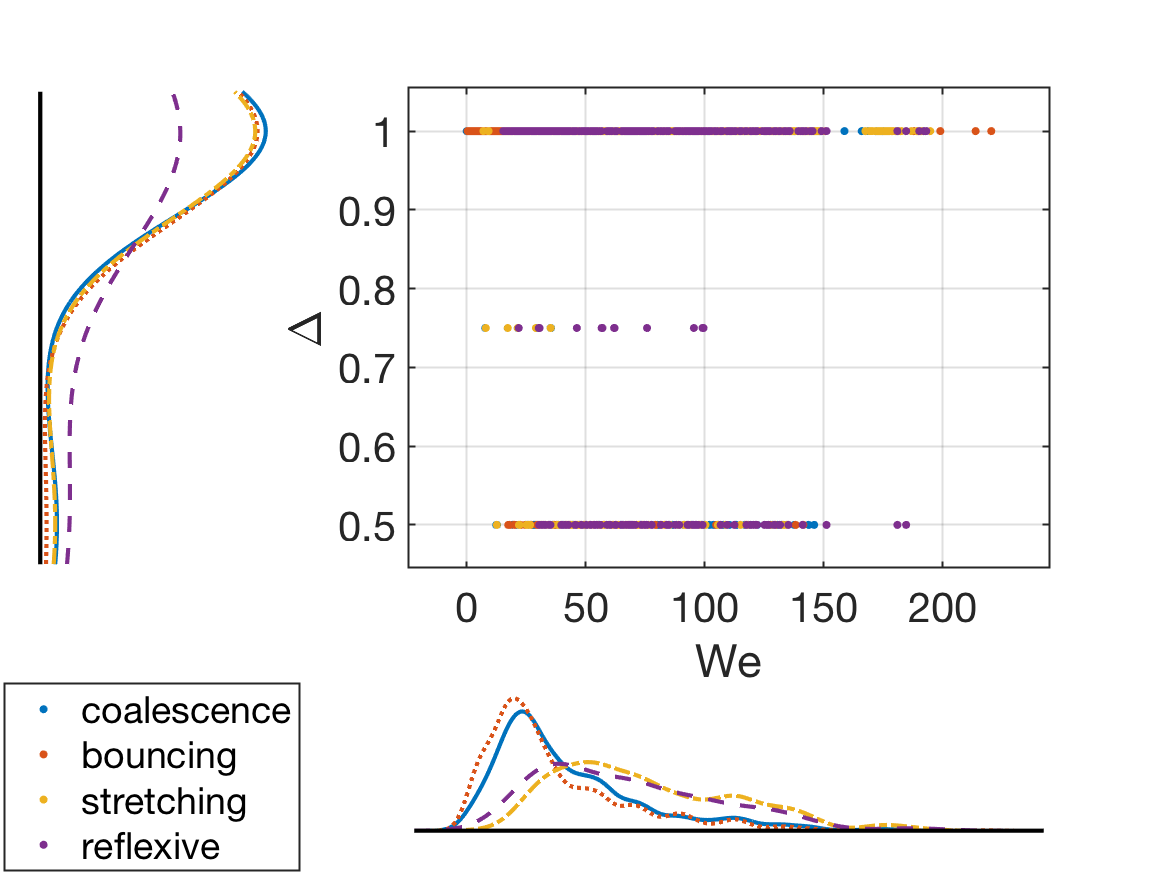}
        \caption{\centering Collision outcomes with respect to $\Delta$ and $We$ (features 1 and 3)}
        \label{fig:data1}
    \end{subfigure}
    ~ 
    \begin{subfigure}[b]{0.4\textwidth}
        \includegraphics[width=\textwidth]{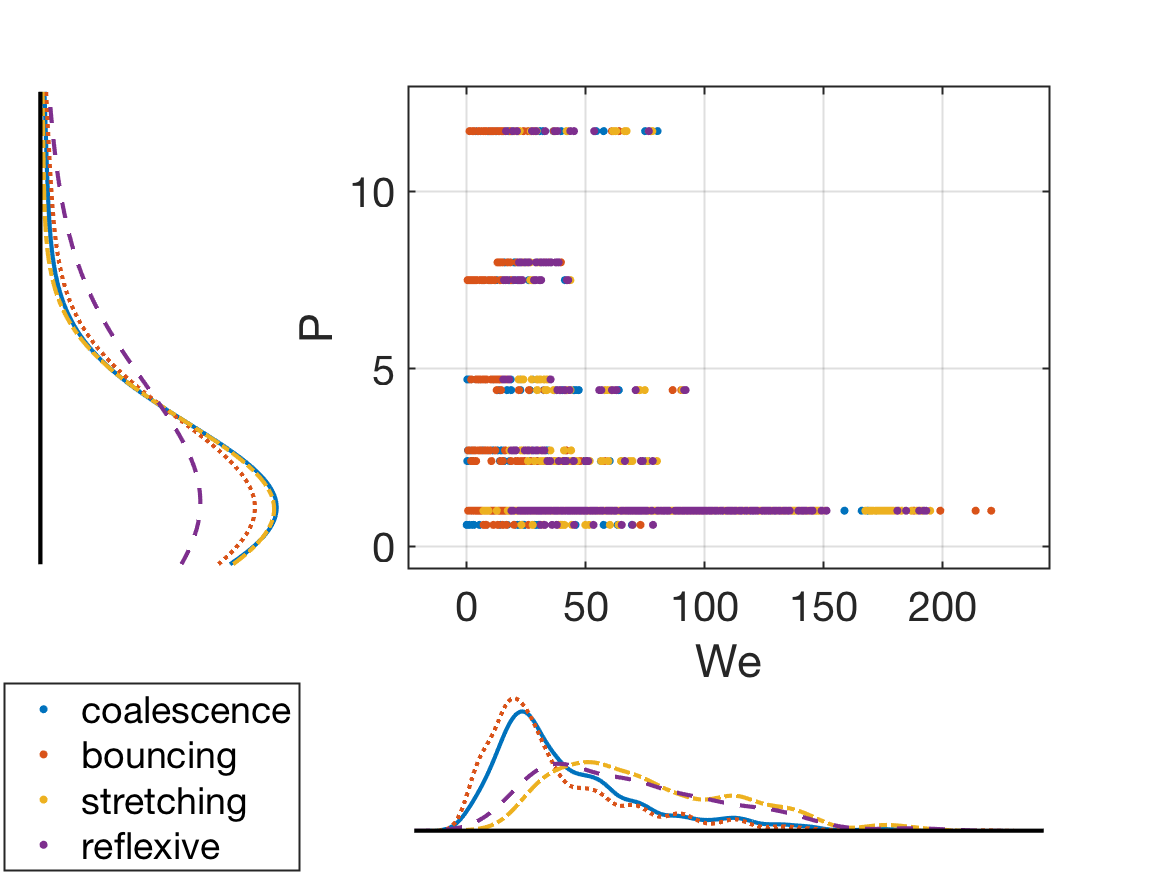}
        \caption{\centering Collision outcomes with respect to $P$ and $We$ (features 1 and 4)}
        \label{fig:data2}
    \end{subfigure}
    ~ 
    \begin{subfigure}[b]{0.4\textwidth}
        \includegraphics[width=\textwidth]{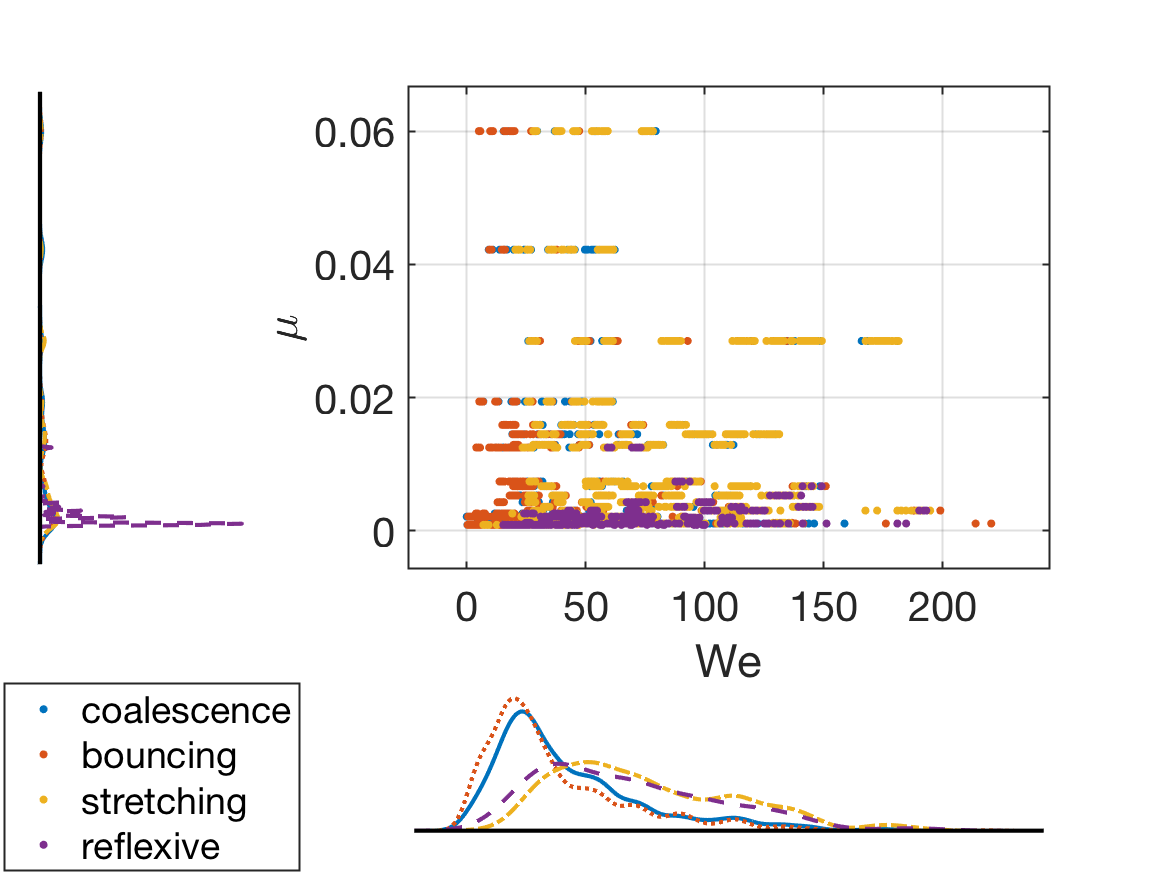}
        \caption{\centering Collision outcomes with respect to $\mu$ and $We$ (features 1 and 5)}
        \label{fig:data3}
    \end{subfigure}
        ~ 
    \begin{subfigure}[b]{0.4\textwidth}
        \includegraphics[width=\textwidth]{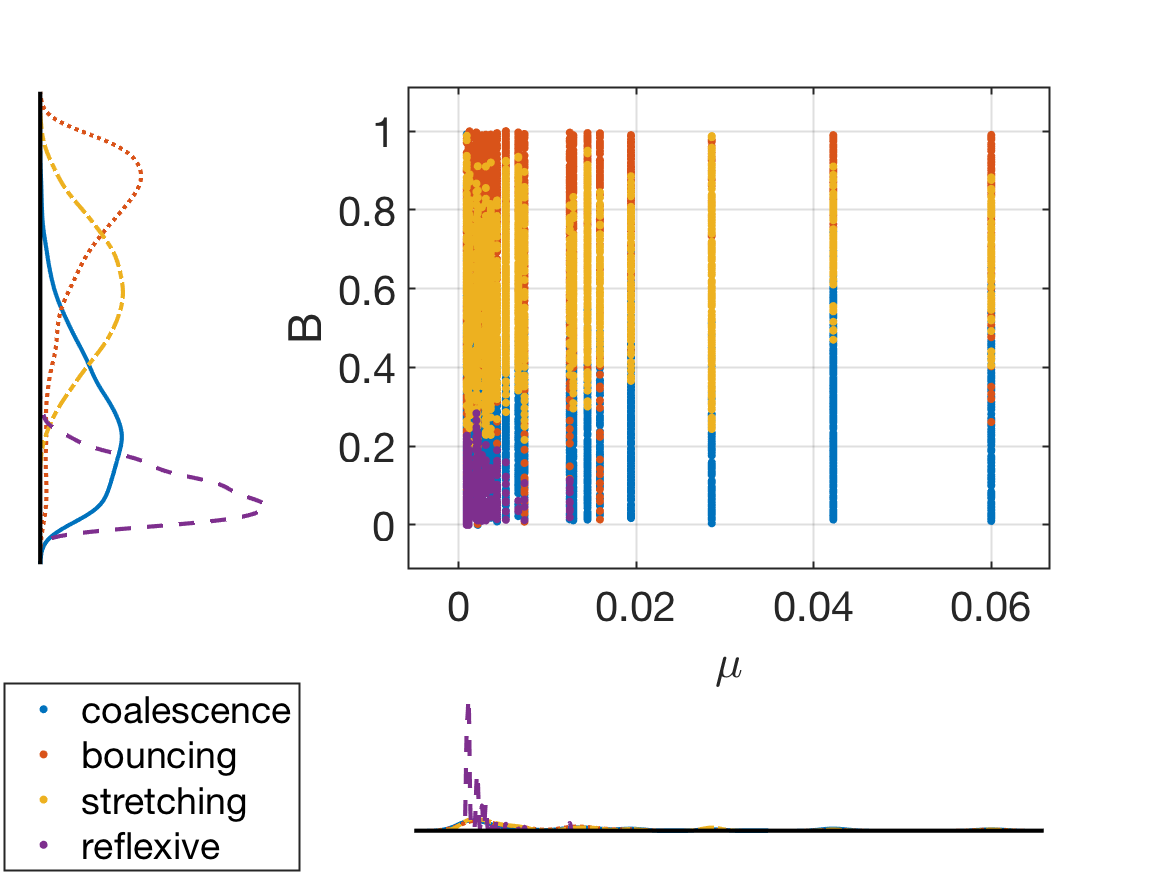}
        \caption{\centering Collision outcomes with respect to $B$ and $\mu$ (features 5 and 2)}
        \label{fig:data4}
    \end{subfigure}
    	\caption{Overall droplet collision dataset (N = 7898, details in Table~\ref{tab:data}), showing variation in collision outcomes with respect to the different features listed in Table~\ref{tab:features}.} \label{fig:animals}
\end{figure}

\subsection{Incorporating domain knowledge}
The 3 decision curves that form the physics based models \cite{munnannur2007new} are given below:

\begin{equation}
\label{eq:c1}
 C_1 = We - \frac{\Delta(1+\Delta^2)(4\Phi - 12)}{\chi_1\left[ \cos\left( \sin^{-1}B\right) \right]}
\end{equation}
\begin{equation}
\label{eq:c2}
 C_2 = We - \frac{3[7(1+\Delta^3)^{2/3} - 4(1+\Delta^2)]\Delta(1+\Delta^3)^2}{\Delta^6 \eta_1 + \eta_2}
\end{equation}
\begin{equation}
\label{eq:c3}
C_3: B - \sqrt{\frac{2.4(\Delta^3- 2.4 \Delta^2 + 2.7\Delta)}{We}}
\end{equation}

These curves divide the feature space or the physical input space into separate regions, and depending on where the datapoint lies with respect to these curves, the physics based models assign the outcome (see Figure~\ref{fig:regime}). Here we incorporate these 3 decision curves as features into our model. $C_1$, $C_2$ and $C_3$ are the additional features formed through combinations of the base features. 

\subsection{Gaps in the data set}
The traditional physics based models only consider 3 features ($\Delta, We$, $B$), therefore, in reporting the data some authors do not report values for the other two features ($P, \mu$) considered here. To fill the gaps I have made educated guesses based on the description of the setup and the fluid used of those features. Furthermore, \citet{qian1997regimes} do not distinguish between classes 3 and 4, and report them as the same. Here, again, I have made educated guesses for their labels based on where the datapoints lie in the regime map (Figure~\ref{fig:regime}).

\section{Machine Learning Modeling}
Scikit-learn~\cite{scikit_learn} and numpy~\cite{harris2020array} are the two main libraries used through this work. All the machine learning model implementations are used directly from scikit-learn. As part of preprocessing, the dataset is shuffled randomly and the feature values are standardized (scaled  to a mean of 0 and standard deviation of 1). Throughout the paper, ten-fold cross validation results have been reported.

\subsection{Hyperparameter tuning}
\label{sec:tuning}
In this section we tune hyperparameters for a few non-linear classifiers. For each of the classifiers, we consider two settings:
\begin{enumerate} 
\item Simply use the 5 base features, i.e., no domain knowledge used
\item Use a total of 8 features (5 base + 3 extra), i.e., leverage domain knowledge
\end{enumerate}

\subsubsection{Decision Trees}
\begin{figure}[h!]
    \centering
    \begin{subfigure}[b]{0.45\textwidth}
        \includegraphics[width=\textwidth]{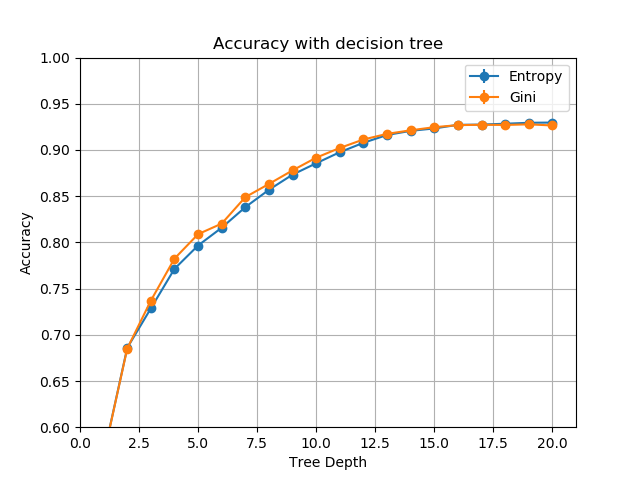}
        \caption{Without domain knowledge (5  features)}
        \label{fig:dt_wo}
    \end{subfigure}
    ~ 
    \begin{subfigure}[b]{0.45\textwidth}
        \includegraphics[width=\textwidth]{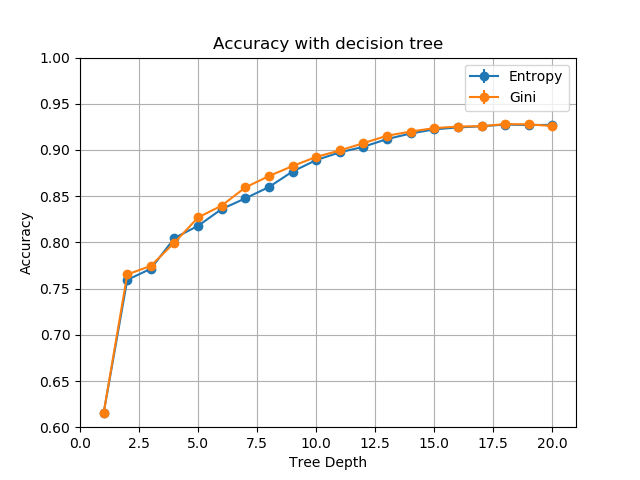}
        \caption{With domain knowledge (8  features)}
        \label{fig:dt_w}
    \end{subfigure}
    \caption{Model prediction accuracy (10-fold cross-validation tests) with decision trees}
    \label{fig:dt}
\end{figure}

\begin{figure}[h!]
    \centering
    \begin{subfigure}[b]{0.45\textwidth}
        \includegraphics[width=\textwidth]{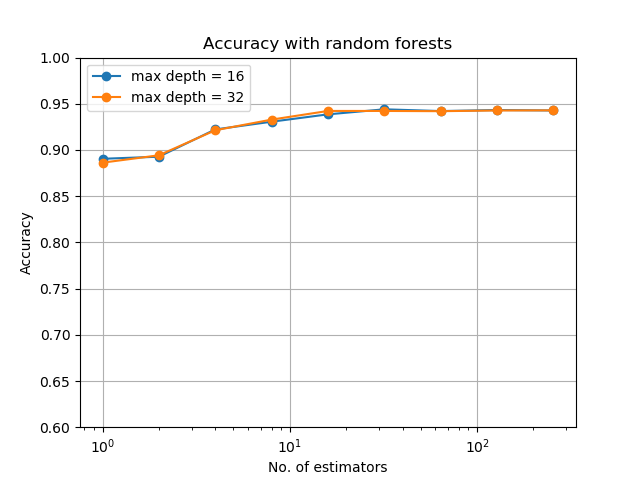}
        \caption{Without domain knowledge (5  features)}
        \label{fig:rf_wo}
    \end{subfigure}
    ~ 
    \begin{subfigure}[b]{0.45\textwidth}
        \includegraphics[width=\textwidth]{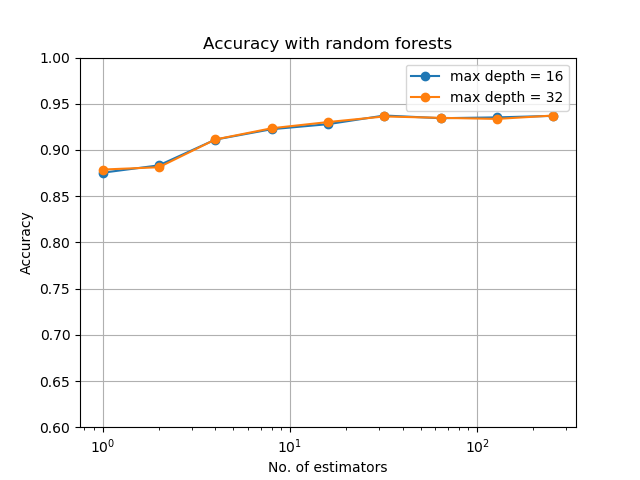}
        \caption{With domain knowledge (8  features)}
        \label{fig:rf_w}
    \end{subfigure}
    \caption{Model prediction accuracy (10-fold cross-validation tests) with random forests}\label{fig:rf}
\end{figure}

Here we test two different splitting criterions over different values of maximum tree depth. A few observations can be made:
\begin{itemize}
\item When tree depth is short (< 5), the 3 additional features significantly improve the prediction accuracy. The extra features play a diminishing role in accuracy as the trees get deeper. The accuracy values for both the cases (Figure~\ref{fig:dt_wo} and Figure~\ref{fig:dt_w}) saturate around 0.93.
\item Choice of splitting criterion (gini impurity vs. entropy) has little effect on accuracy across all tests. The accuracy from both criteria remains roughly the same across tree depths and in 5 features vs. 8.
\item Accuracy saturates beyond max tree depth of around 20. This value remains similar with 5 input features and 8 input features. This is confirmed in experiments with random forests as well (see Figure~\ref{fig:rf}).
\end{itemize}

\subsubsection{Random Forests}

Here we test accuracy vs. the number of estimators used. We do this for two different values of max tree depth. This is repeated with both cases, one with 5 features and one with the 3 extra features. A few observations can be made from these tests:
\begin{itemize}
\item The accuracy values are fairly high (close to 0.95 for high number of estimators).
\item The extra features have no impact on the accuracy. Even for a small number of estimators. This makes sense considering the results seen in Figure~\ref{fig:dt}. The accuracy of decision trees was not affected by the 3 additional features beyond a tree depth of 16. Here we consider tree depths of 16 and 32, therefore, even with 1 estimator, the accuracy values are not affected by the additional features.
\item The maximum tree depth values (beyond 16) has no impact on accuracy of the random forests.
\item The accuracy value is practically saturated beyond 30 estimators for all cases.
\end{itemize}

\subsubsection{K-Nearest Neighbors}
\begin{figure}[h!]
    \centering
    \begin{subfigure}[b]{0.45\textwidth}
        \includegraphics[width=\textwidth]{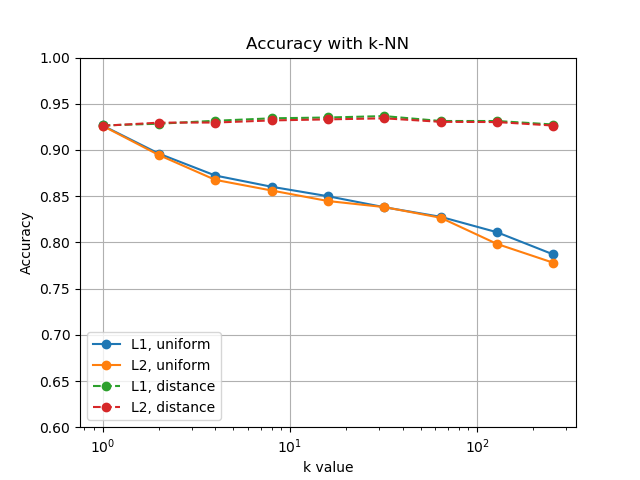}
        \caption{Without domain knowledge (5  features)}
        \label{fig:gull}
    \end{subfigure}
    ~ 
    \begin{subfigure}[b]{0.45\textwidth}
        \includegraphics[width=\textwidth]{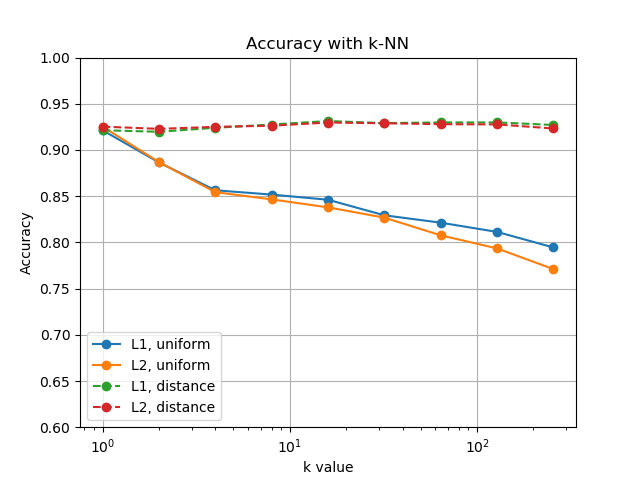}
        \caption{With domain knowledge (8  features)}
        \label{fig:tiger}
    \end{subfigure}
    \caption{Model prediction accuracy (10-fold cross-validation tests) with k-nn}\label{fig:knn}
\end{figure}

Here we experiment with different distance metrics ($L_1$ and $L_2$) and different label computation options (uniform and distance weighted averaging). These experiments are repeated for both cases, with the 5 base features and with all the 8 features. Results are presented in Figure~\ref{fig:knn}.

Here are a few observations based on results in Figure~\ref{fig:knn}:
\begin{itemize}
\item The 3 additional features provide an insignificant improvement in the prediction accuracy.
\item Label calculation method directly impacts accuracy.
\subitem With $k=1$, we have a single neighbor, therefore the distance weighted label coincides with the direct computation. 
\subitem As the $k$ value increases the uniformly weighted label accuracy drops quickly. $k=1$ is the optimal value for uniform weighing. This makes sense when we consider the data in Figure~\ref{fig:data0}. There is no gap between the different classes, moreover, there is some mixing of the classes as well. Therefore higher $k$ values pollute the labels.
\subitem This issue is remedied in the distance weighted labels, where closer points are weighted higher. Here we observe a small increase in accuracy with increasing $k$ values up to $k=30$.
\item Finally, the the $L_1$ and $L_2$ distance metrics have little impact on the accuracy.
\end{itemize}

\subsubsection{Neural Networks}
\begin{figure}[h!]
    \centering
    \begin{subfigure}[b]{0.45\textwidth}
        \includegraphics[width=\textwidth]{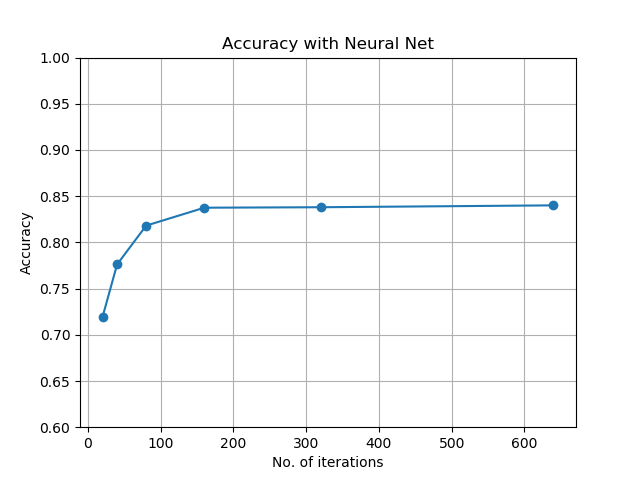}
        \caption{Without domain knowledge (5  features)}
        \label{fig:gull}
    \end{subfigure}
    ~ 
    \begin{subfigure}[b]{0.45\textwidth}
        \includegraphics[width=\textwidth]{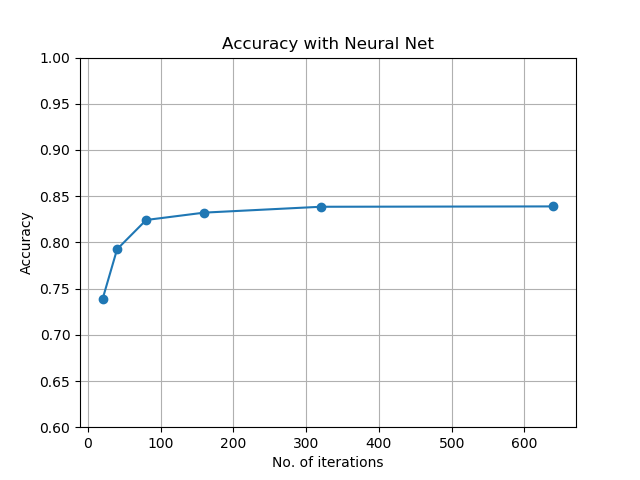}
        \caption{With domain knowledge (8  features)}
        \label{fig:tiger}
    \end{subfigure}
    \caption{Model prediction accuracy (10-fold cross-validation tests) with Neural Networks}\label{fig:nn}
\end{figure}

Here we consider the accuracy of neural network classifiers as a function of number of training epochs/iterations. We tested different neural network architectures (different number of hidden layers and neurons per layer) but they performed similarly with 2 or more hidden layers with more than 10 neurons each. In Figure \ref{fig:nn} we present results with a 5 hidden layers, first four have 40 neurons and the last one has 8 neurons. Once again we see that the additional features have little impact on the accuracy. Also, the accuracy saturates beyond 300 iterations.

\subsection{Accuracy with tuned models}
\begin{figure}[h!]
    \centering
    \begin{subfigure}[b]{0.45\textwidth}
        \includegraphics[width=\textwidth]{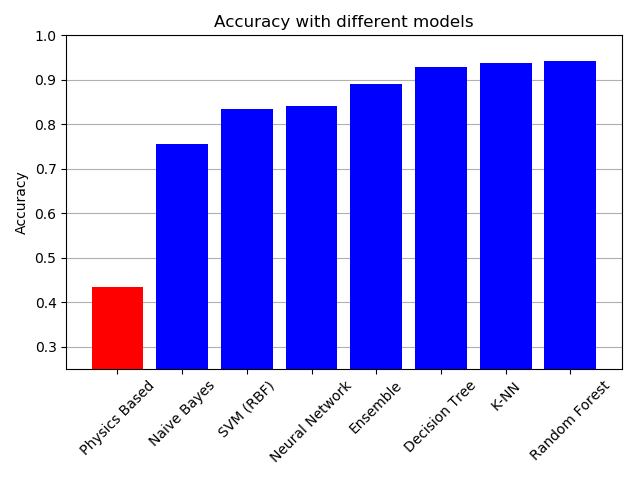}
        \caption{Without domain knowledge (5  features)}
        \label{fig:gull}
    \end{subfigure}
    ~ 
    \begin{subfigure}[b]{0.45\textwidth}
        \includegraphics[width=\textwidth]{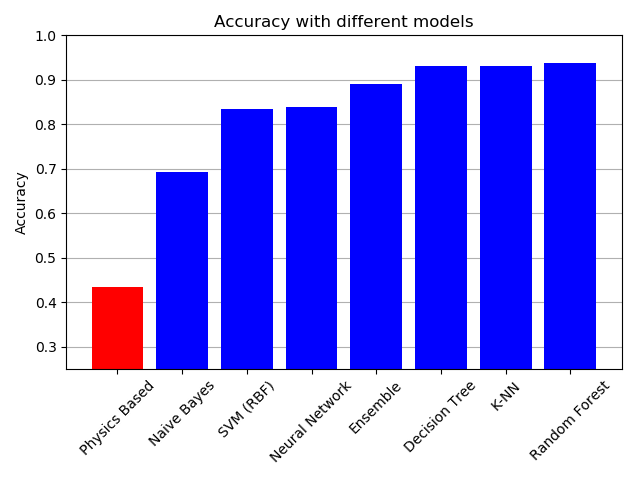}
        \caption{With domain knowledge (8  features)}
        \label{fig:tiger}
    \end{subfigure}
    \caption{Model prediction accuracy (10-fold cross-validation tests) across different models}\label{fig:tuned}
\end{figure}

In this section we compare the accuracy of different machine learning models (with tuned hyperparameters based on prior section) for the two cases,  one with the 5 base features and the other with the additional features. The results are presented in Figure~\ref{fig:tuned}. The accuracy of the state of the art physics based models~\cite{agarwal2019computational} is also presented (in red) as a baseline. The `ensemble' method refers to a majority vote of the random forest, neural network, SVM and k-NN predictions. 

A few comments and observations regarding the results:
\begin{itemize}
\item Even the simplest ML models provide a significant improvement over the current state of the art physics based models. This is mainly due to the low model complexity of the physics based models.
\item Little difference in the accuracy in the two cases, with and without domain knowledge. This is consistent with our observations in Figures~\ref{fig:dt}--\ref{fig:nn}. The naive bayes classifier actually suffers with the additional features.
\item The random forest classifier performs the best on this dataset. Although it is only marginally better than the much simpler decision tree and k-NN classifiers.
\item This ensemble classifier does not boost the accuracy. An ensemble is expected to improve accuracy when the different classifiers exploit different aspects of the dataset.
\end{itemize}

\begin{figure}[h]
	\centering
	\includegraphics[width=0.6\textwidth]{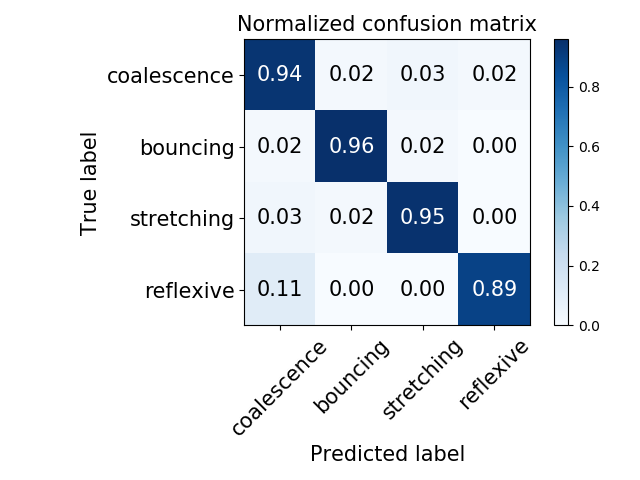}
	\caption{Normalized confusion matrix for the random forest classifier.}
	\label{fig:conf}
\end{figure}

The normalized confusion matrix, showing the predicted labels vs. the true lables, for the random forest classifier is presented in Figure~\ref{fig:conf} as a sample. The random forest classifier is picked because it has the highest accuracy. The accuracy is fairly high for the first 3 classes. The highest error is when the true label is `reflexive' and it gets classified as `coalescence'. More training data near this boundary can help boost the model accuracy further. This can possibly guide further experiments.

\subsection{Why did domain knowledge not help?}
A key question that arises from the experiments presented so far is why the domain knowledge in the form of the additional features did not help the performance of any of the models. To answer this we consider an experiment with increasing size of the dataset. The two simple classifiers that perform well (see Figure~\ref{fig:tuned}), decision tree and k-NN, are considered for this test. For both the classifiers we consider training with 5 features (dotted lines) and 8 features (solid lines).

\begin{figure}[h!]
    \centering
    \begin{subfigure}[b]{0.475\textwidth}
        \includegraphics[width=\textwidth]{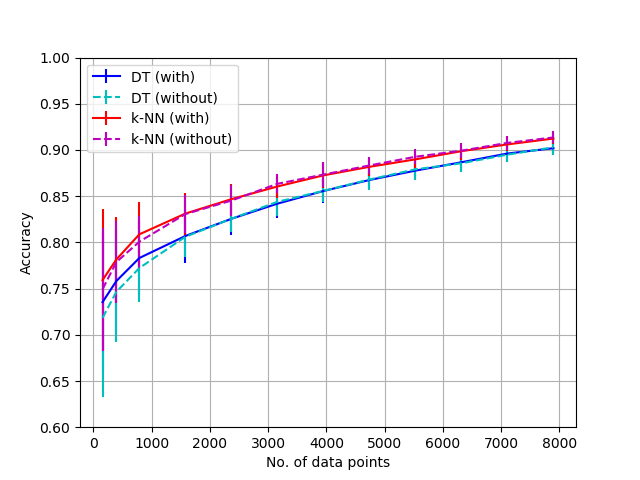}
        \caption{Over the whole range of dataset sizes}
        \label{fig:acc_vs_size_a}
    \end{subfigure}
    ~ 
    \begin{subfigure}[b]{0.475\textwidth}
        \includegraphics[width=\textwidth]{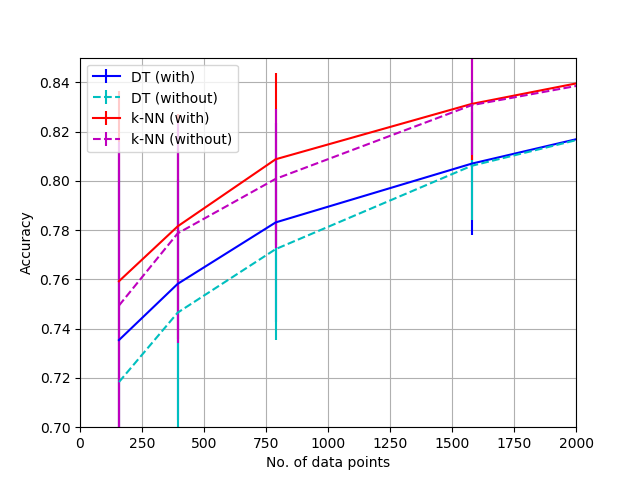}
        \caption{For small dataset sizes (zoomed into Figure~\ref{fig:acc_vs_size_a})}
        \label{fig:acc_vs_size_b}
    \end{subfigure}
    \caption{Model accuracy (10-fold cross-validation tests) with different models as a function of number of training data points.}\label{fig:acc_vs_size}
\end{figure}

A few observations can be made:
\begin{itemize}
\item The accuracy has not saturated with respect to increasing number of data points. More data is expected to boost accuracy of our models further.
\item Domain knowledge does not help boost accuracy when the training dataset is large (2000+). However, for a smaller number of training data points, there is an improvement when domain knowledge is used. This is highlighted in Figure~\ref{fig:acc_vs_size_b}, where we zoom in on the relevant region.
\subitem The improvement (albeit small) in performance by addition of the extra features suggests that when there is a small amount of training data the models are unable to search the entire hypothesis space. The addition of the relevant features allows a better accuracy even when the hypothesis space is not completely spanned.
\subitem This improvement vanishes as the number of training data points increases because the models are able to better search the hypothesis space with additional data.
\end{itemize}

\subsection{Robustness to noise}
A key metric to consider for the current application is the robustness to noise. As described in Table~\ref{tab:data}, the training data has been extracted from different sources going as far back as 1990. Certain experimental errors (up to 10\%) have been reported in literature and some variations can be expected across the different datasets. Therefore, we must account for noise in our input data. To do this we run experiments where we add a varying degree of noise to the data.

\begin{figure}[h]
	\centering
	\includegraphics[width=0.7\textwidth]{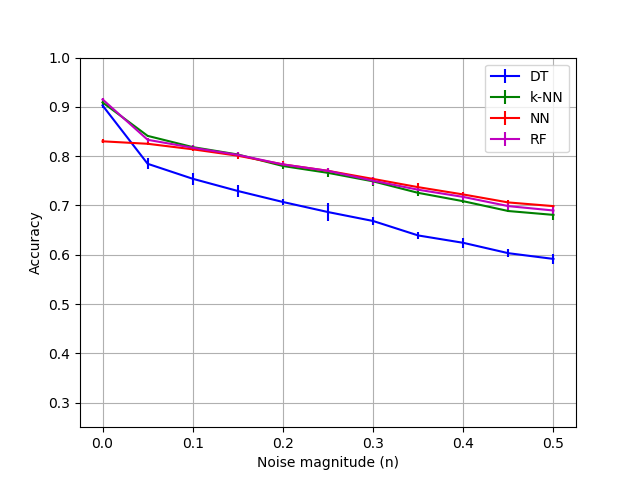}
	\caption{Accuracy of different classifiers as a function of noise in the data. The $n$ value denotes the standard deviation in the Gaussian noise. }
	\label{fig:noise}
\end{figure}

Gaussian noise is added to the scaled feature vectors. The noisy data is given as 
\begin{equation}
X(i, j) = X(i, j) + N(i, j),
\end{equation}
where $N \sim \mathcal{N}(0, n)$, $i$ is the instance index and $j$ is the feature number. The value of $n$ is represented on the x-axis in Figure~\ref{fig:noise}. Four different classifiers are compared here, decision tree (DT), k-NN, neural network (NN) and random forests (RF). 

Two key things emerge from this test:
\begin{itemize}
\item The decision tree classifier is the least robust to noise. There is a sharp drop in accuracy initially up to 5\% noise. The other 3 models are more robust to input noise.
\item The neural network classifier, while starting from a lowest accuracy is the most robust to input noise. The neural network performs even better than the random forest for high levels of noise.
\end{itemize}

\section{Conclusions}

A major improvement is seen over the current state of the art physics based models ($\approx 43\%$) with all ML classifiers studied here ($> 90\%$). The accuracy is more than doubled. The best classifiers for the current dataset based on three key metrics are:
\begin{itemize}
\item Accuracy: Random forests
\item Interpretability: Decision tree
\item Robustness: Neural Networks
\end{itemize}
The k-NN classifier performs decently well on all three criteria.

Furthermore, the model accuracy has not saturated with the current size of the training dataset (see Figure~\ref{fig:acc_vs_size}). In the confusion matrix (Figure~\ref{fig:conf}) the highest error is seen when the true label is `reflexive' and it gets classified as `coalescence'. More training data near this boundary can help boost the model accuracy further.

Droplet collision remains a relevant fluid-dynamics problem for several industries (e.g., automotive, food processing). This paper demonstrates that ML based classifiers can significantly improve the prediction of droplet collision outcomes. Moreover, other than simply improving the model accuracy, this work can be leveraged to provide insights into the fluid mechanics of the problem itself. Interpreting the present ML based models, mainly the decision tree, can lead to further insights about the dependence of the droplet collision phenomena on the different features.

\bibliography{ml_droplet_collision}
\end{document}